\def\year{2021}\relax
\documentclass[letterpaper]{article}
\usepackage{aaai21}
\usepackage{times}
\usepackage{helvet}
\usepackage{courier}
\usepackage[hyphens]{url}
\usepackage{graphicx}
\usepackage{natbib}
\usepackage{caption}
\usepackage{booktabs}
\usepackage{multirow}
\usepackage{amsmath}
\usepackage{amsfonts}
\usepackage{amssymb}
\usepackage{pifont}
\usepackage[pagebackref=true,breaklinks=true,letterpaper=true,colorlinks,bookmarks=false]{hyperref}
\hypersetup{colorlinks=true, linkcolor=blue, anchorcolor=blue, citecolor=blue, filecolor=blue, urlcolor=blue}

\title{Few-shot Font Generation with Localized Style Representations and Factorization}
\author {
    Song Park\textsuperscript{\rm 1, \thanks{Equal contribution. Work done at NAVER CLOVA.}}~
    Sanghyuk Chun\textsuperscript{\rm 2, 3 \footnotemark[1]}~
    Junbum Cha\textsuperscript{\rm 3}~~
    Bado Lee\textsuperscript{\rm 3}~~
    Hyunjung Shim\textsuperscript{\rm 1, \thanks{Hyunjung Shim is a corresponding author.}} \\
}
\affiliations {
    \textsuperscript{\rm 1} School of Integrated Technology, Yonsei University, \quad
    \textsuperscript{\rm 2} NAVER AI LAB, \quad
    \textsuperscript{\rm 3} NAVER CLOVA
}

\usepackage{color}
\usepackage{xspace}
\newcommand{\eg}{\textit{e.g.}\xspace}
\newcommand{\ie}{\textit{i.e.}\xspace}
\newcommand{\ours}{LF-Font\xspace}
\newcommand{\methodname}{few-shot font generation with localized style representations and factorization (LF-Font)\xspace}
\definecolor{darkergreen}{RGB}{21, 152, 56}
\definecolor{red2}{RGB}{252, 54, 65}
\newcommand{\yesmark}{\textcolor{darkergreen}{\ding{52}}}
\newcommand{\nomark}{\textcolor{red2}{\ding{56}}}

\begin{document}

\maketitle
\begin{abstract}
Automatic few-shot font generation is a practical and widely studied problem because manual designs are expensive and sensitive to the expertise of designers. Existing few-shot font generation methods aim to learn to disentangle the style and content element from a few reference glyphs, and mainly focus on a universal style representation for each font style. However, such approach limits the model in representing diverse local styles, and thus makes it unsuitable to the most complicated letter system, \eg, Chinese, whose characters consist of a varying number of components (often called ``radical'') with a highly complex structure. In this paper, we propose a novel font generation method by learning localized styles, namely component-wise style representations, instead of universal styles. The proposed style representations enable us to synthesize complex local details in text designs. However, learning component-wise styles solely from reference glyphs is infeasible in the few-shot font generation scenario, when a target script has a large number of components, \eg, over 200 for Chinese. To reduce the number of reference glyphs, we simplify component-wise styles by a product of component factor and style factor, inspired by low-rank matrix factorization. Thanks to the combination of strong representation and a compact factorization strategy, our method shows remarkably better few-shot font generation results (with only 8 reference glyph images) than other state-of-the-arts, without utilizing strong locality supervision, \eg, location of each component, skeleton, or strokes. The source code is available at \url{https://github.com/clovaai/lffont}.
\end{abstract}

\section{Introduction}
\label{sec:intro}
Text is a critical resource taking a considerable portion of the information on the web. Thus, text design is essential to improve the quality of services and user experiences. However, font design is labor-intensive and heavily depends on the expertise of designers, especially for glyph-rich scripts such as Chinese. For this reason, various font generation methods have been investigated to address a few-shot font generation problem, which uses only a few reference font images for automatically generating all the glyphs.

In this paper, we tackle a few-shot font generation problem; generating a new font library with very few references, \eg, $8$. Without additional training procedure (\eg, finetune the model on the reference characters), our goal is to generate high quality, diverse styles in the few-shot font generation scenario. Our few-shot generation scenario consists of training and generation stages. During model training, we rely on many paired data which is easily accessible by public font libraries. On the other hand, at the generation stage, we use only few-shot examples as unseen style references without additional model finetuning. This scenario is particularly effective when the target style glyphs are expensive to collect, \eg, historical handwriting, but we have a large database for existing fonts, or computing resources are limited to run additional finetuning, \eg, on mobile devices. A popular strategy to tackle the same problem is to separate style and content representations from the given glyph images~\cite{sun2018_ijcai_savae,zhang2018_cvpr_emd,gao2019agisnet,srivatsan2019_emnlp_deepfactorization}. These methods generate a full font library by combining the target style representation and the source content representations.

However, previous few-shot font generation methods learn a {\it universal} style representation for each style, limited in representing diverse local styles. It is particularly problematic when generating fonts for glyph-rich scripts, \eg, Chinese, Korean, and Thai. Every Chinese character consists of a varying number of components (often called ``radical'') with a highly complex structure. This property induces the visual quality of a Chinese character to be highly sensitive to local damage or a distinctive local component-wise style.

The same issue was also pointed out by~\citet{cha2020dmfont,cha2020dmfontw} in that many previous methods often fail to transfer unseen styles for the few-shot Korean and Thai generation. To alleviate this problem, \citet{cha2020dmfont} propose dual-memory architecture, named DM-Font. DM-font extracts component-wise local features for all components at once, and then save them into two types of memory. Despite its notable generation quality, DM-Font is restricted to {\it complete compositional scripts}, such as Korean and Thai~\cite{cha2020dmfont,cha2020dmfontw}. While each Korean or Thai character can be decomposed into the fixed number of components and positions, more complex script like Chinese can be decomposed into varying components and positions. As a result, DM-Font fails to disentangle complex glyph structures and diverse local styles in the Chinese generation task, as shown in our experiments. Furthermore, DM-Font requires that all components are shown in the reference set at least once to construct their memories. These drawbacks make DM-Font not applicable to generate Chinese characters, consisting of hundreds of components, with a few references.

In this paper, we propose a novel \methodname. \ours learn to disentangle complex glyph structures and {\it localized} style representations, instead of {\it universal} style representations. Owing to powerful representations, \ours can capture local details in rich text design, thus successfully handle Chinese compositionality. Our disentanglement strategy preserves highly complex glyph structures, while DM-Font~\cite{cha2020dmfont} frequently loses the content information of complex Chinese characters. Consequently, our method shows remarkably better stylization performance than {\it universal} style encoding methods~\cite{sun2018_ijcai_savae, zhang2018_cvpr_emd, gao2019agisnet}. 

We define the localized style representation as a character-wise style feature which considers both a complex character structure and local styles. Instead of handling the large amount of characters in the glyph-rich script, we denote {\it the localized style representation} as a combination of component-wise local style representations (\S~\ref{subsec:localized_style_representation}). However, this strategy can have an inherent limitation; the reference set must cover the whole component set to construct the complete font library. It is infeasible when a target script has a large number of components, \eg, over 200 for Chinese. To solve this issue, we introduce {\it factorization modules}, which factorizes a localized style feature to a component factor and a style factor (\S~\ref{subsec:factorization}). Consequently, our method can generate the whole vocabulary without having the entire components in the reference style, or utilizing strong locality supervision, \eg, location of each component, skeleton, or strokes.

We demonstrate the effectiveness of the proposed \ours on the Chinese few-shot font generation scenario when the number of references is extremely small (namely, $8$) (\S~\ref{sec:experiment}). Our method outperforms five state-of-the-art few-shot font generation methods with various evaluation metrics, with a significant gap. Careful ablation studies on our design choice shows that the proposed localized style representation and factorization module are an effective choice to tackle our target problem successively.

\section{Related Works}
\label{sec:relwork}

\paragraph{Font generation as image-to-image (I2I) translation.}
I2I translation~\cite{isola2017_cvpr_pix2pix, zhu2017_iccv_cyclegan} aims to learn a mapping between source and target domains while preserving the contents in the source domain, \eg, day to night. Recent I2I translation methods are extended to learn a mapping between multiple diverse domains~\cite{stargan, liu2018unified, yu2019multi, starganv2}, \ie, multi-domain translation, thus can be naturally adopted into the font generation problem. For example, \citet{zi2zi} attempted to solve the font generation task via paired I2I translation by mapping a fixed ``source'' font to the target font.

\paragraph{Few-shot font generation.}
The few-shot font generation task aims to generate new glyphs with very few numbers of style references without additional finetuning. The mainstream of few-shot font generation attempts to disentangle content and style representations as style transfer methods~\cite{gatys2016neuralstyle, adain, wct, deepphotostyle, photowct, wct2}, but specialized to font generation tasks. For example, AGIS-Net~\cite{gao2019agisnet} proposes the font-specialized local texture discriminator and the local texture refinement loss. Unlike other methods, DM-Font~\cite{cha2020dmfont} disassembles glyphs to stylized components and reassembles them to new glyphs by utilizing strong compositionality prior, rather than disentangles content and style.

Despite notable improvement over past years, previous few-shot font generation methods have significant drawbacks, such as infeasible to generate complex glyph-rich scripts~\cite{azadi2018mcgan}, failing to capture the local diverse styles~\cite{sun2018_ijcai_savae,zhang2018_cvpr_emd,gao2019agisnet,liu2019funit,srivatsan2019_emnlp_deepfactorization}, or losing the complex content structures~\cite{cha2020dmfont,cha2020dmfontw}. This paper proposes a novel few-shot font generation method that disentangles complex local glyph structure and diverse local styles, resulting in high visual quality of the generated samples for complex glyph-rich scripts, \eg, Chinese.

\paragraph{Other Chinese font generation methods.}
Although we only focus on the few-shot font generation problem, there are a several papers address the Chinese font generation task with numerous references or additional finetuning. SCFont~\cite{jiang2019_aaai_scfont} and ChiroGAN~\cite{gao2020_aaai_chirogan} extract a skeleton or a stroke from the source glyphs and translate it to the target style. They require a large number of references for generating glyphs with a new style, \eg, $775$~\cite{jiang2019_aaai_scfont}. Instead of expensive skeleton or stroke annotations, another approach~\cite{sun2018_ijcai_savae,huang2020_eccv_rdgan,wu2020calligan,cha2020dmfont} utilizes the compositionality to reduce the expensive search space in the character space to smaller component space. RD-GAN~\cite{huang2020_eccv_rdgan} aims to generate unseen characters in the fixed style, not feasible to our few-shot font generation scenario. CalliGAN~\cite{wu2020calligan} encodes the styles by one-hot vectors; thus it requires additional finetuning for making unseen style during the training. ChiroGAN~\cite{gao2020_aaai_chirogan} aims to solve unpaired font generation tasks as unpaired image-to-image translation tasks~\cite{zhu2017_iccv_cyclegan}. However, in our scenario, glyph images can be easily rendered from an existing font library, building a paired training dataset is cheap and does not limit the practical usage.

\section{Few-shot Font Generation with Localized Style Representations and Factorization}
\label{sec:method}
We propose a novel few-shot font generation framework, \methodname, having strong representational power even with a very few reference glyphs, by introducing localized style features and factorize modules. 

\subsection{Problem definition}
\label{subsec:probdef}

\begin{figure}[t]
    \centering
    \includegraphics[width=0.8\linewidth]{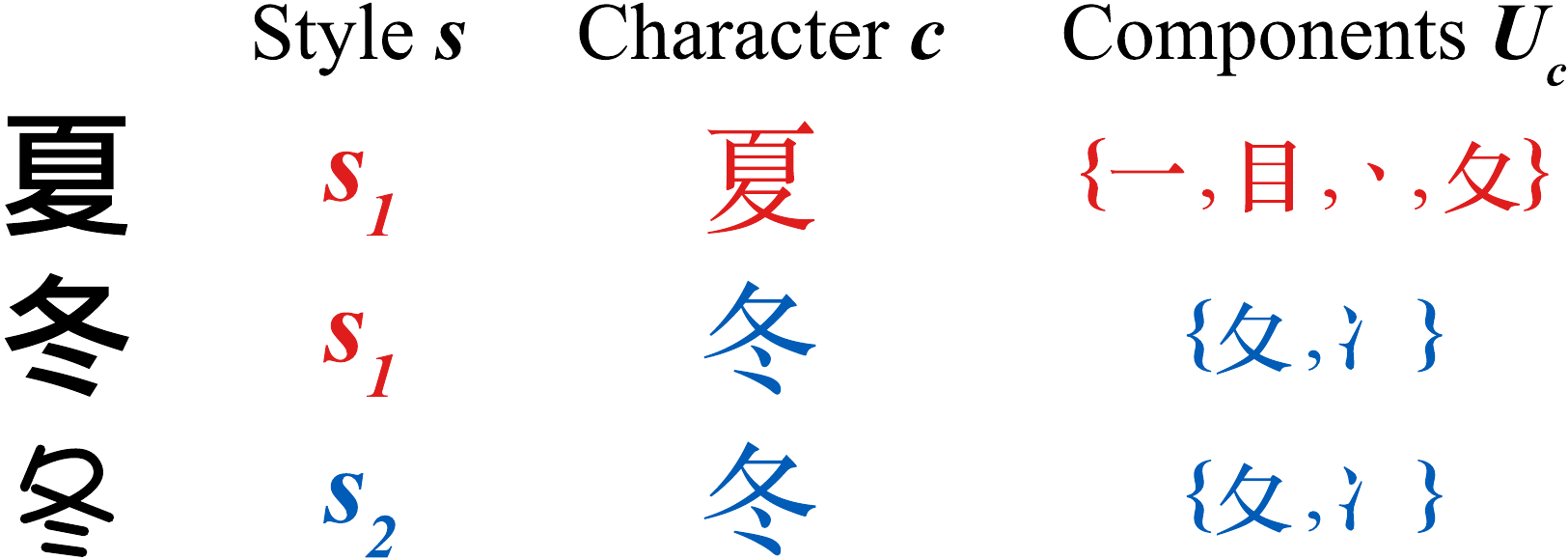}%
    \caption{\small {\bf Annotation examples.} The character label $c$, the style label $s \in \{s_1, s_2\}$ and the component label set $U_c$ is shown.}%
    \label{fig:example_annotation}
\end{figure}

We define three annotations for a glyph image $x$: the style label $s \in \mathcal S$, the character label $c \in \mathcal C$, and the component labels $U_c = [u^c_1, \ldots, u^c_m]$, where $m$ is the number of components in character $c$. Here, each character $c$ can be decomposed into components $U_c$ by the pre-defined decomposition rule as Figure~\ref{fig:example_annotation}. In our Chinese experiments, the number of the styles $|\mathcal S| = 482$, the number of the characters $|\mathcal C| = 19,514$, and the number of the components $|U| = 371$. In other words, all $19,514$ characters can be represented by the combination of $371$ components. Note that our problem definition is not limited to the Chinese, but easily extended to other languages as shown in Appendix.

\begin{figure*}[t]
    \centering
    \includegraphics[width=\linewidth]{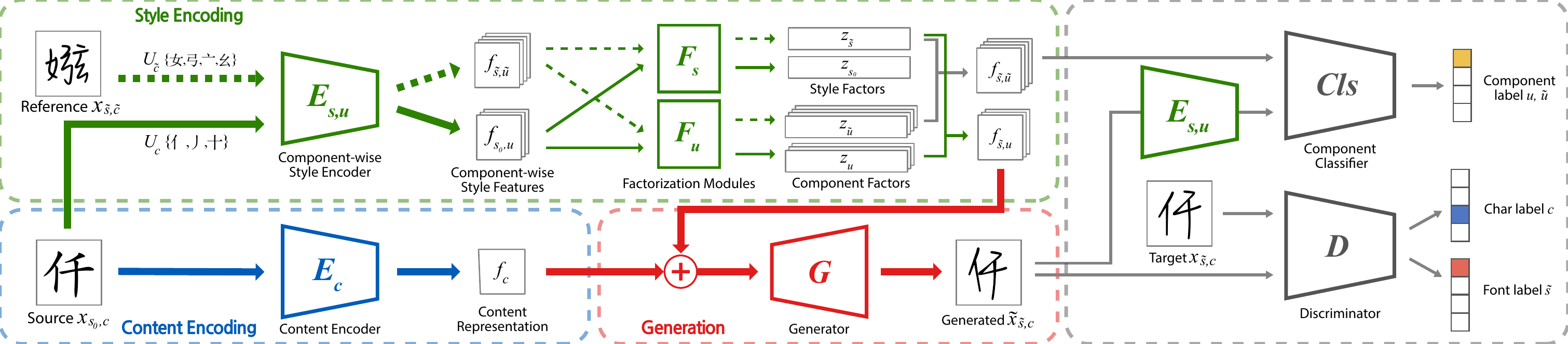}%
    \caption{\small {\bf Overview of \ours.} \ours consists of four parts; the content-encoding $E_c$, the style-encoding $E_{s,u}, F_s, F_u$, the generation $G$, and the shared modules $D, Cls$ for training. $E_c$ encodes the source glyph to the content representation $f_c$. In our style encoding stage, the source image (solid line) and reference images (dashed line) are encoded to component-wise style features $f_{s,u}$, and further factorized into style and component factors $z_s, z_u$. The extracted style and component factors are combined to the character-wise style representation $f_{s,c}$ of the target glyph. The generator $G$ synthesizes the target glyph from the content feature $f_c$ and the localized style feature $f_{s,c}$. The discriminator and the component classifier are employed for training.
 }%
    \label{fig:architecture}
\end{figure*}

\begin{figure}[h]
    \centering
    \includegraphics[width=0.8\linewidth]{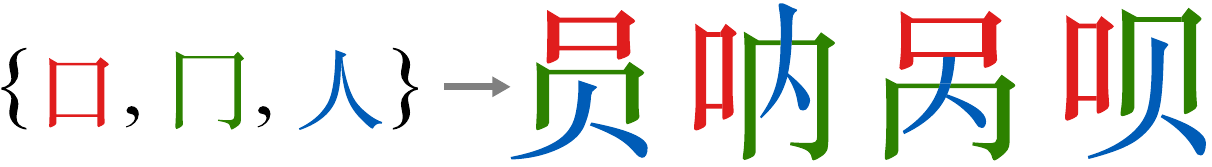}%
    \caption{\small {\bf Characters from the same component set.} Examples to show that a component set is mapped to diverse characters.}%
    \label{fig:twins}
\end{figure}

The goal of few-shot font generation task is to generate a glyph $x_{\widetilde s, c}$ with unseen target style $\widetilde s$ for all $c \in \mathcal C$ with very few number of references $x_{\widetilde s, \widetilde c} \in \mathcal X_r$, \eg, $| \mathcal X_r | = 8$. A common framework for few-shot font generation is to learn a generator $G$, which takes the style representation $f_{\widetilde s} \in \mathbb R^d$ from $\mathcal X_r$ and the content representation $f_c \in \mathbb R^d$ as inputs, and synthesize a glpyh $x$ having reference styles $\widetilde s$ but representing a source character $c$. It is formulated by developing the generator $G$ and encoders $E_s$ and $E_c$ for extracting style and content representations, respectively as follows:

\begin{align}
\label{eq:fewshotfontgeneral}
\begin{split}
    &x_{\widetilde s, c} = G( f_{\widetilde s}, f_c ), \\
    f_{\widetilde s} = &E_s(\mathcal X_r) ~\text{and}~ f_c = E_c( x_{s_0, c} ),
\end{split}
\end{align}
where $s_0$ is the source style label.

\subsection{Localized style representations}
\label{subsec:localized_style_representation}
Previous methods assume that the style representation $f_s$ is universal for each style $s$, uniquely determined over all characters. However, the universal style assumption can overlook complex local styles, resulting in poor performances for unseen styles, as pointed by~\cite{cha2020dmfont}. Here, we design the style encoder $E_s$ to encode character-wise style. This strategy is useful when a style is defined very locally and diversely as Chinese characters. However, the huge vocabulary size of Chinese script ($|\mathcal C| > 20,000$) makes it impossible to exploit all character-wise styles.

Instead of handling all character-wise styles, we first represent the character as a combination of multiple components, and develop the component-wise styles to minimize the redundancy in character-level representations. For that, we utilize the component set $U_c$ instead of the character label $c$, where $|U| \ll |\mathcal C|$. We extract a component-wise style feature $f_{s,u}(x, u) = E_{s,u}(x, u) \in \mathbb R^d$ from a reference glyph image $x$ and a component label $u \in U_c$ by introducing a component-wise style encoder $E_{s,u}$. Then, we compute {\it the character-aware localized style feature} $f_{s,c}$ by taking the summation over component-wise features $f_{s,u}$. Now, we can rewrite Eq~\eqref{eq:fewshotfontgeneral} with the proposed character-aware localized style features as follows:

\begin{align}
\label{eq:fewshotfontdm}
\begin{split}
    x (\widetilde s, c) &= G(f_{\widetilde s,c}, f_c ), \quad f_c = E_c( x_{s_0, c} ), \\
    f_{\widetilde s,c} &= \sum_{u \in U_c} f_{\widetilde s,u} = \sum_{u \in U_c} E_{s,u}(x_{\widetilde s, \widetilde c_u}, u),
\end{split}
\end{align}
where $x_{\widetilde s, \widetilde c_u}$ is a glyph image from reference set $\mathcal X_r$ whose character is $\widetilde c_u$, which contains component $u$. However, the minimum required size of $\mathcal X_r$ is too large for Chinese because a total number of component set $\mathcal U$ in Chinese is still too large, \eg, $229$.

\subsection{Completing missing localized style representations by factorization modules}
\label{subsec:factorization}
In our scenario, only partial components are observable from the reference set, while the other components are not accessible by $E_{s,u}$. Hence, the localized style feature $f_{s,c}$ for a style $s$ and a character $c$ with unseen components cannot be computed, and $G$ therefore, cannot generate a glyph with $c$. In other words, the few-shot font generation is not achievable if the reference glyphs cannot cover the whole component set ($|U| = 371$). To tackle the problem, we formulate the few-shot font generation problem as a reconstruction problem; given observations with a few style-component pairs, we aim to reconstruct the missing style-component pairs. Inspired from classical matrix completion approaches~\cite{candes2009exact,cai2010singular}, we decompose the component-wise style feature $f_{s,u} \in \mathbb R^d$ into two factors: a component factor $z_u \in \mathbb R^{k \times d}$ and a style factor $z_s \in \mathbb R^{k \times d}$, where $k$ is the dimension of factors. Formally, we decompose $f_{s,u}$ into $z_s$ and $z_u$ as follows:
\begin{equation}
\label{eq:factorization}
    f_{s,u} = \boldsymbol{1}^\top (z_s \odot z_u),
\end{equation}
where $\odot$ is an element-wise matrix multiplication, and $\boldsymbol{1} \in \mathbb R^k$ is an all-ones vector. Eq~\eqref{eq:factorization} can be interpreted as the element-wise matrix factorization of $f_{s,u}$.
In practice, we extract the style factor $z_s$ from the reference set and combine them with the component factor $z_u$ from the source glyph to reconstruct a component-wise style feature $f_{s,u}$ for the given source character $c$.
Note that \citet{tenenbaum2000separating, srivatsan2019_emnlp_deepfactorization} also use a factorization strategy to font generation, but they directly apply the factorization to the complex glyph space, \ie, each element is an image, while \ours factorizes the localized style features into the style and the content factors.

Traditional matrix completion methods require heavy computations and memory consumption. For example, expensive convex optimization~\cite{candes2009exact} or alternative algorithm~\cite{cai2010singular}, are infeasible in our scenario: repeatedly apply matrix factorization $d$ times to obtain a $d$ dimensional feature $f_{s,u}$. Instead, we propose an style and component factorization modules $F_s$ and $F_u$ which extracts factors $z_s, z_u \in \mathbb R^{k \times d}$ from the given feature $f_{s,u} \in \mathbb R^d$ as follows:
\begin{equation}
\label{eq:factorization_modules}
    z_s = F_s (f_{s,u}; W, b), \quad z_u = F_u (f_{s,u}; W, b).
\end{equation}
We use a linear weight $W = [w_1; \ldots; w_k] \in \mathbb R^{k \times d}$ and a bias $b \in \mathbb R^k$ as a factorization module, where each factor is computed by $z = [w_1 \odot f_{s,u} + b_1; \ldots; w_k \odot f_{s,u} + b_k]$.

Note that solely employing the factorization modules, \ie, Eq~\eqref{eq:factorization_modules}, does not guarantee that factors with the same style (or component) from different glyphs have identical values. Thus, we train the factorization modules $F_s$ and $F_u$ by minimizing the consistency loss $\mathcal L_{consist}$ as follows:
\begin{align}
\begin{split}
\label{eq:loss_factorization_consist}
    \mathcal L_{consist} = \sum_{s \in \mathcal S} \sum_{u \in \mathcal U} &\| F_s(f_{s,u}) - \mu_s \|_2^2 + \| F_u(f_{s,u}) - \mu_u \|_2^2, \\
    \mu_s = \frac{1}{|\mathcal U|}\sum_{u \in \mathcal U} F_s&(f_{s,u}), \quad \mu_u = \frac{1}{|\mathcal S|}\sum_{s \in \mathcal S} F_u(f_{s,u}).
\end{split}
\end{align}

After training $F$, we can extract $z_s$ from even a random single reference glyph. Furthermore, by combining $z_s$ with the content factor $z_u$ from the known source glyph, we can reconstruct the localized style feature $f_{s, c} = \sum_{u \in U_c} f_{s, u}$ even for the unseen component $u$ in the reference set.

\subsection{Generation}
\label{subsec:generation}
Once \ours is trained with many paired training samples, it is able to generate any unseen style fonts with only a few references by extracting the style factor $z_{\widetilde s}$ from the reference glyphs, and by extracting $z_u$ and $f_c$ from the known source glyphs. Then, we combine $z_c$ and $z_{\widetilde s}$ for generating the localized style feature $f_{\widetilde s, u}$ as described in \S\ref{subsec:factorization}. Finally, we generate a glyph $x$ using Eq \eqref{eq:fewshotfontdm}. Formally, \ours consists of three sub-modules as illustrated in Figure~\ref{fig:architecture}. We describe the details of each sub-module below.

\paragraph{Style encoding.}
\ours encodes the localized style representation $f_{s,c}$ by encoding the component-wise features $f_{s,u}$ as formulated in Eqs~\eqref{eq:fewshotfontdm},~\eqref{eq:factorization} and~\eqref{eq:factorization_modules}. There are three main modules in this stage: the component-wise style encoder $E_{s,u}$, the style and content factorization modules $F_s$ and $F_c$. $E_{s,u}$ is simply defined by a conditional encoder where a component label $u$ is used for the condition label, and encodes a glyph image $x$ into several component-wise style features $f_{s,u}$. 

A component-wise style feature $f_{s,u}$ is factorized into the style factor $z_s$ and the component factor $z_u$ with factorization modules $F_s$ and $F_u$, respectively. We combine the style factor $z_{\widetilde s}$ from the reference glyphs and the component factor $z_u$ from the source glyph to reconstruct re-stylized the component-wise feature $f_{\widetilde s,u}$. If there are more than one reference sample, we take the average over the style factors, extracted from each reference glyph, to compute $z_{\widetilde s}$.

\paragraph{Content encoding.}
Although our proposed style encoding strategy effectively captures the local component information, it requires guidance on the complex global structure (\eg, relative locations of components) of each character, because a component set can be mapped to many characters -- See Figure~\ref{fig:twins}. We employ the content encoder $E_c$ to capture the complex global structural information of the source glyph. It facilitates to generate the target glyph while preserving complex structural information without any strong localization supervision of the source glyph.

\paragraph{Generation.}
Finally, the generator $G$ produces the target glyph $\widetilde x_{\widetilde s, c}$ by combining the localized style representations $f_{\widetilde s,c}$ from the style encoding and the global complex structural representation $f_{c}$ from the encoding.

\subsection{Training}
\label{subsec:training}
Given the source glyph $x$ and the references $\mathcal X_r$ having the target style $s$, \ours learns the style encoder $E_{s,u}$, the content encoder $E_c$, the factorization modules $F_s, F_u$, and the generator $G$ for generating a glyph $\widetilde x$. We fix the source style $s_0$ during training and optimize the model parameters with diverse reference styles using the following losses.

\paragraph{Adversarial loss.}
As we strive to generate a plausible glyph in terms of both style and content, we employ a multi-head conditional discriminator for style label $s$ and character label $c$. The hinge GAN loss~\cite{zhang2019sagan} is used.

\begin{align}
\begin{split}
\label{eq:loss_adv}
    \mathcal L_{adv}^D = &-\mathbb E_{(x, s, c) \sim p_{data}} \max\left(0, -1 + D_{s,c}(x) \right) \\
    &- \mathbb E_{(\widetilde x, s, c) \sim p_{gen}} \max\left(0, -1 - D_{s,c}(\widetilde x) \right)\\
    \mathcal L_{adv}^G = &-\mathbb E_{(\widetilde x, s, c) \sim p_{gen}} D_{s,c}(\widetilde x).
\end{split}
\end{align}

\paragraph{L1 loss and feature matching loss.}
These objectives enforce the generated glyph $\widetilde x$ to reconstruct the ground truth glyph $x$ in pixel-level and feature-level.

\begin{align}
\begin{split}
\label{eq:loss_recon}
    \mathcal L_{l1} &= \mathbb E_{(x, s, c) \sim p_{data}} \left[ \| x - \widetilde x \|_1 \right],\\
    \mathcal L_{feat} &= \mathbb E_{(x, s, c) \sim p_{data}} \left[\sum_{l=1}^L\| D_f^{(l)}(x) - D_f^{(l)}( \widetilde x ) \|_1 \right]
\end{split}
\end{align}
where $L$ is the number of layers in the discriminator $D$ and $D_f^{(l)} (x)$ is the intermediate feature in the $l$-th layer of $D$.

\paragraph{Component-classification loss.}
We employ additional component-wise classifier $Cls$ which classifies the component label $u$ of the given component-wise style feature $f_{s,u}$. We optimize the cross entropy loss (CE) as follows:
\begin{equation}
\label{eq:loss_cls}
    \mathcal L_{cls} = \sum_{\widetilde u \in U_{\widetilde c}} \text{CE}(Cls(f_{s, \widetilde u}), \widetilde u) + \sum_{u \in U_c} \text{CE}(Cls(f_{s,u}), u),
\end{equation}
where $f_{s,\widetilde u}$ and $f_{s,u}$ are extracted from the reference glyph $x_{s, \widetilde c}$, and the generated glyph $\widetilde x_{s, c}$.

\paragraph{Full objective.}
Finally, we optimize \ours by the following full objective function:

\begin{align}
\begin{split}
    &\min_{\substack{E_c, E_{s,u}, G, \\F_s, F_u, Cls}}\max_D~ \mathcal L_{adv (font)} + \mathcal L_{adv (char)} + \lambda_{L1} \mathcal L_{L1} \\
    &\qquad+ \lambda_{feat} \mathcal L_{feat} + \lambda_{cls} \mathcal L_{cls} + \lambda_{consist} \mathcal L_{consist},
\end{split}
\end{align}
where $\lambda_{L1}, \lambda_{feat}, \lambda_{cls}, \lambda_{rep}$ are hyperparameters for controlling the effect of each objective. We set $\lambda_{L1} = 1.0$ and $\lambda_{feat} = \lambda_{cls} = \lambda_{rep} = 0.1$ throughout all the experiments.

\paragraph{Training details.}
We optimize our model with Adam optimizer~\cite{adam}. For stable training, we first train the model without factorization modules as Eq~\eqref{eq:fewshotfontdm}. Here, the model is trained to generate a target glyph from the component-wise style features $f_{s,u}$ directly extracted from the reference set $\mathcal X_r$. We construct a mini-batch with pairs of a reference set and a target glyph. To build each pair, we randomly select a style from the training style set and construct a reference set and a target glyph, where the components of the target glyph belong to the components in the reference set, but the target glyph is not in $\mathcal X_r$. After enough iterations, we add the factorization modules to the model and jointly train all modules. In this phase, the reference set is changed to have diverse styles and the target style is randomly chosen from the reference styles. More details of our method are described in Appendix.

\begin{table}[b]
\small
\centering
\begin{tabular}{@{}lccc@{}}
\toprule
         & Localized & Contents & Restricted \\ 
         &  style?   & encoder? & to generate \\ \midrule
SA-VAE   & \nomark          & \nomark                & unseen chars (train)\\
EMD      & \nomark          & \yesmark               & \\
AGIS-Net & \nomark          & \yesmark               & \\
FUNIT    & \nomark          & \yesmark               & \\
DM-Font  & \yesmark         & \nomark                & unseen components (refs.) \\ \midrule
Ours     & \yesmark         & \yesmark               &                                                                                            \\ \bottomrule
\end{tabular}
\caption{\small {\bf Comparison of \ours with other methods.} We show the taxonomy of few-shot font generation by the localized style and the content encoder. Note that SA-VAE cannot generate unseen characters during the training, and DM-Font is unable to synthesis a glyph whose component is not observable in the reference glyphs.}
\label{table:comparion_methods_desc}
\end{table}
\begin{table*}[ht!]
\centering
\small
\setlength{\tabcolsep}{4pt}
\begin{tabular}{@{}cclccccccccc@{}}
\toprule
                  &&                   & LPIPS $\downarrow$ &  & Acc (S) $\uparrow$ & Acc (C) $\uparrow$ & Acc (Hmean) $\uparrow$ &  & FID (S) $\downarrow$ & FID (C) $\downarrow$ & FID (Hmean) $\downarrow$ \\ \midrule
\parbox[t]{2mm}{\multirow{6}{*}{\rotatebox[origin=c]{90}{Seen chars}}}   && SA-VAE (IJCAI'18) & 0.310 &  & 0.2       & 41.0        & 0.3         &  & 231.8     & 66.7        & 103.6       \\
                  && EMD (CVPR'18)     & 0.248 &  & 11.9      & 63.7        & 20.1        &  & 148.1     & 25.7        & 43.8        \\
                  && AGIS-Net (TOG'19) & 0.182 &  & 34.0      & \textbf{99.8}  & 50.7     &  & 79.8      & 4.0         & 7.7         \\
                  && FUNIT (ICCV'19)   & 0.217 &  & 39.0      & 97.1        & 55.7        &  & 58.5      & 3.6         & 6.8         \\
                  && DM-Font (ECCV'20) & 0.275 &  & 10.2      & 72.4        & 17.9        &  & 151.8     & 8.0         & 15.2        \\ %
                  && \ours (proposed)              & \textbf{0.169} &  & \textbf{75.6}      & 96.6       & \textbf{84.8}        &  & \textbf{40.4}      & \textbf{2.6}         & \textbf{4.9}         \\ \midrule
\parbox[t]{2mm}{\multirow{5}{*}{\rotatebox[origin=c]{90}{Unseen chars}}} && EMD (CVPR'18)     & 0.250 &  & 11.6      & 64.0        & 19.7        &  & 151.7     & 41.4        & 65.0        \\
                  && AGIS-Net (TOG'19) & 0.189 &  & 33.3      & \textbf{99.7}        & 49.9        &  & 85.4      & 10.0        & 18.0        \\
                  && FUNIT (ICCV'19)   & 0.216 &  & 38.0      & 96.8        & 54.5        &  & 63.2      & 12.3        & 20.6        \\
                  && DM-Font (ECCV'20) & 0.284 &  & 11.1      & 53.0        & 18.4        &  & 153.4     & 26.5        & 45.2        \\ %
                  && \ours (proposed)              & \textbf{0.169} &  & \textbf{72.8}      & 97.1        & \textbf{83.2}        &  & \textbf{44.5}      & \textbf{8.7}         & \textbf{14.6}       \\ \bottomrule
\end{tabular}
\caption{\small {\bf Performance comparison on few-shot font generation scenario.} Six few-shot font generation methods are compared with eight reference glyphs. LPIPS shows a perceptual similarity between the ground truth and the generated glyphs. We also report accuracy and FID measured by style-aware (S) and content-aware (C) classifiers. The harmonic mean (Hmean) of style- and content-aware metrics shows the overall visual quality of the generated glyphs. All numbers are average of $50$ runs with different reference glyphs.}
\label{table:main_fewshot}
\end{table*}

\begin{figure*}[t]
    \centering
    \includegraphics[width=\linewidth]{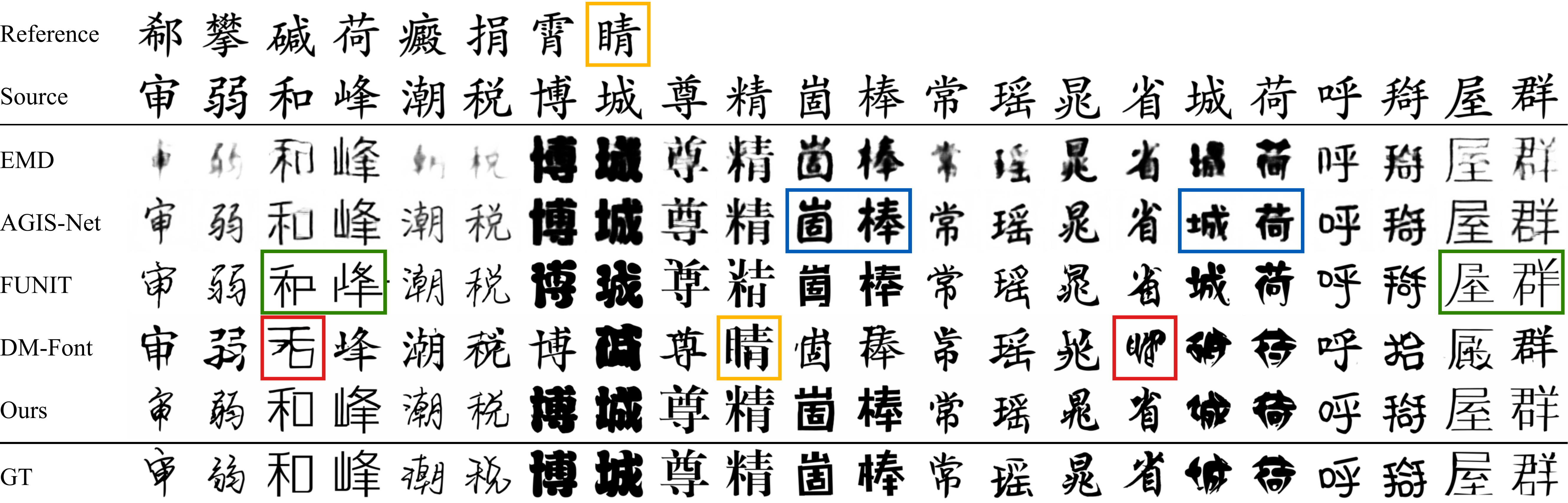}%
    \caption{\small {\bf Generated samples.} We show characters in the reference set (refer the character only, not style), source images, generated samples of \ours and five comparison methods, and the target glyphs (see GT). The reference images in each style are provided in Appendix. We also highlight samples which show the apparent limitation of each method by the colored boxes. Each color denotes the different failure cases discussed in \S~\ref{subsec:main_results}.}%
    \label{fig:main_few}
\end{figure*}

\section{Experiments}
\label{sec:experiment}
This section shows the comparison results of \ours and previous methods in the Chinese few-shot font generation (Korean generation results are also shown in Appendix). Extensive analysis shows that our design choice successfully deals with the few-shot font generation task. We also provide ablation studies on the effects of objective functions, size of the reference set, and factor size $k$ in Appendix.

\subsection{Datasets and evaluation metrics}
\label{subsec:datasets}
We collect public $482$ Chinese fonts from the web. The dataset has a total of $19,514$ characters (each font has a varying number of characters and it is $6,654$ characters on average), which can be decomposed by $371$ components. We sample $467$ fonts corresponding to $19,234$ characters for training, and the remaining unseen $15$ fonts are used for the evaluation. The models are separately evaluated with $2,615$ {\it seen characters} and $280$ {\it unseen characters} to measure the generalizability to the unseen characters.

We evaluate the visual quality of generated glyphs using various metrics. To measure how faithful the generated glyphs match their ground truths, LPIPS~\cite{zhang2018_cvpr_lpips} with ImageNet pre-trained VGG-16 is used. LPIPS is popularly used for assessing the similarity between two images by considering the perceptual similarity.

We further assess the visual quality of generated glyphs in two aspects; content-preserving and style-adaptation aspects as \citet{cha2020dmfont}. We train two classifiers, each to distinguish the style or content labels of the test dataset. Note that we train the evaluators independently from our generation models, and the character and font labels for the evaluation have no overlap with training labels. ResNet-50~\cite{he2016_cvpr_resnet} is employed for the backbone architecture.
Comparing to photorealistic images, glyph images are highly sensitive to the local damage or a distinctive local component-wise information. We optimize evaluation classifiers by CutMix augmentation~\cite{yun2019cutmix}, which let a model learn localizable and robust features~\cite{chun2019icmlw}, and AdamP optimizer~\cite{heo2020adamp}. More details are in Appendix. We report the accuracies of the generated glyphs by {\it style-aware} and {\it content-aware} models, respectively. We also use each classifier as a feature extractor and compute Frechét inception distance (FID)~\cite{heusel2017_nips_ttur_fid}. In the experiments, we denote metrics computed by content and style classifiers as {\it content-aware} and {\it style-aware}, respectively.

\subsection{Comparison methods}
\label{subsec:comparison_methods}

We compare our model with five state-of-the-art few-shot font generation methods. For the sake of understanding the similarity or dissimilarity between methods, we categorize them by whether or not they explicitly model style representations or content representations as Table~\ref{table:comparion_methods_desc}.

SA-VAE~\cite{sun2018_ijcai_savae} extracts a universal style feature and utilizes a content code from the character classifier instead of the content encoder. This method cannot synthesize the characters unseen during training.

EMD~\cite{zhang2018_cvpr_emd}, AGIS-Net~\cite{gao2019agisnet}, and FUNIT~\cite{liu2019funit} employ the content encoder but their style representation is universal for the given style. For FUNIT, we use the modified FUNIT for the font task as \citet{cha2020dmfont,cha2020dmfontw}. We empirically show that this universal style representation strategy fails to capture diverse styles, even incorporating specialized modifications, \eg, the local texture discriminator, and the local texture refinement loss for AGIS-Net.

DM-Font~\cite{cha2020dmfont} would be the most direct competitor to \ours. Both DM-Font and \ours utilize the component-wise style features to capture the local details. However, DM-Font is restricted to generate a glyph whose component is not in the reference set because it uses the learned codebook for each component instead of the content encoder. Since DM-Font affords to generate neither Chinese characters nor glyphs with unseen components, we use the source style to extract local features for substituting the component-wise features for the unseen component.
The modification details are described in Appendix.

\subsection{Experimental results}
\label{subsec:main_results}
\paragraph{Quantitative evaluation.} We evaluate the visual quality of the generated images by six models with eight reference glyphs per style. To avoid randomness by the reference selection, we repeat the experiments $50$ times with different reference characters. A font generation method is required to satisfy two contradictory task objectives: it should preserve contents and stylize well. As an extreme failure case, it performs an identity mapping, which will show the perfect content preserving score but it will show zero style transfer score. Hence, we report harmonic mean of content and style scores to probe whether a method can satisfy both objectives well. Table~\ref{table:main_fewshot} shows that our method outperforms previous state-of-the-arts with significant gaps, \eg, 28.7pp higher harmonic mean accuracy than FUNIT, and 3.4 lower harmonic mean FID than AGIS-Net for the unseen characters. Our method particularly outperforms other methods in style-aware benchmarks while the content-aware benchmarks are not much damaged. For example, FUNIT and AGIS-Net show comparable performance in content-aware benchmarks to \ours, but they show far lower performances than \ours in style-aware benchmarks. In other words, FUNIT and AGIS-Net only focus on content preserving, while fail in good stylization.

\paragraph{Qualitative evaluation.} We also compare generated samples by the methods qualitatively in Figure~\ref{fig:main_few}. For the reference style, please to refer the font style in GT and Appendix. We observe that AGIS-Net often drops local details such as serif-ness, varying thickness (blue boxes). The green boxes show that FUNIT overly relies on the structure of source images. Thus, FUNIT tends to destroy the local structures in generated glyphs when the source and the target glyphs' overall structure differ a lot. We argue that the universal style representation strategy by AGIS-Net and FUNIT causes the problems. We further provide extensive analysis of the style representations in the latter section.

We observe that DM-Font frequently fails to generate correct characters. For example, as the red boxes, DM-Font often generates a glyph whose relative component locations are muddled. Another example is in the yellow boxes; DM-Font generates glyphs with the wrong component, observable in the references. We conjecture that the absence of the content encoder makes DM-Font suffer from the complex structures of glyphs. In the latter section, we show that the content encoder is critical to capture the complex structures.

Compared to others, \ours generates the most plausible results that preserve the local details of each component and global structure of characters of target styles.

\begin{figure}[t]
    \centering
    \includegraphics[width=1.0\linewidth]{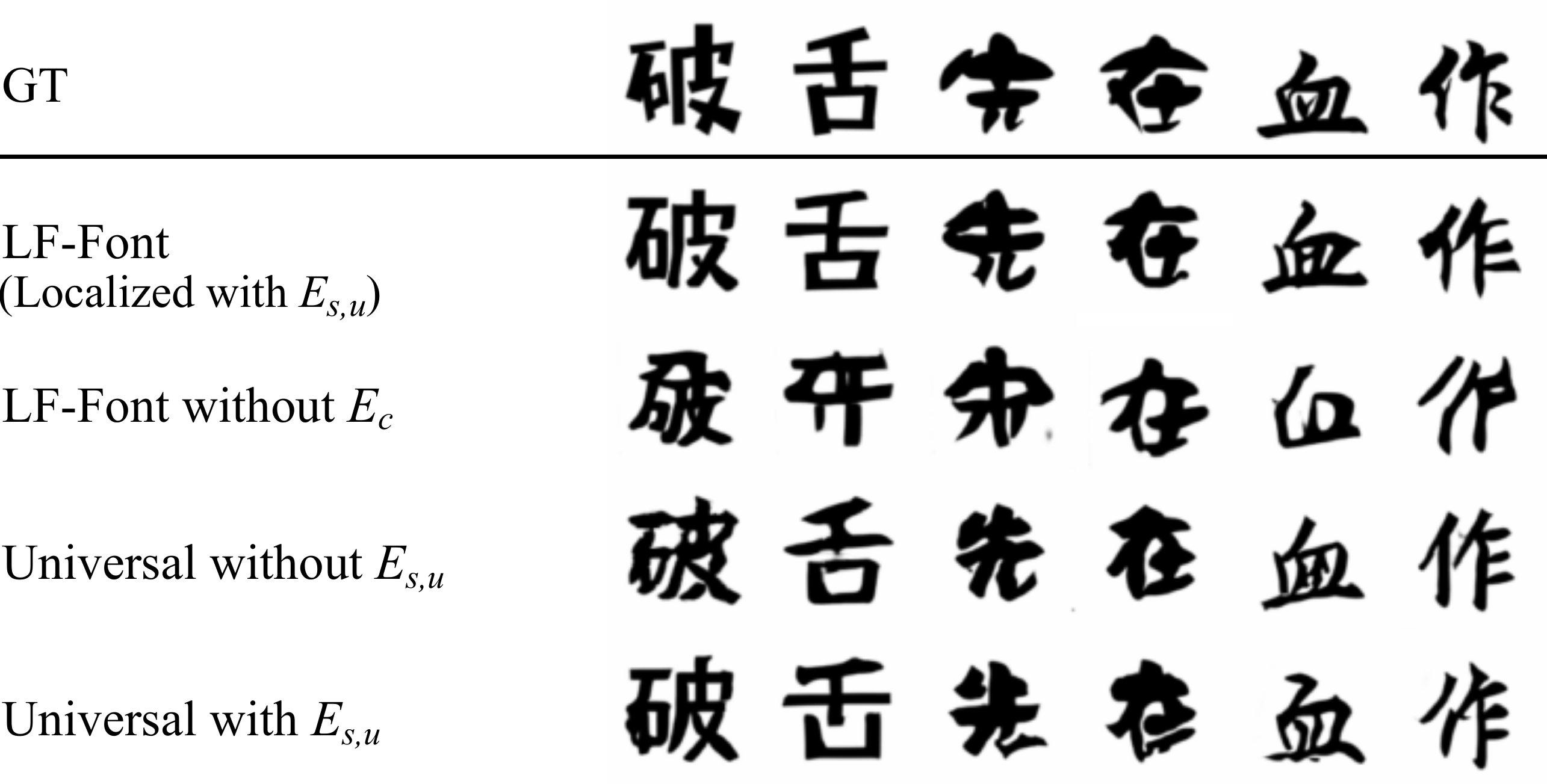}%
    \caption{\small {\bf Visual samples of style and content module analysis.} Visual samples in Table~\ref{table:abl_content} and Table~\ref{table:abl_style} are shown.}%
    \label{fig:example_ablation}
\end{figure}

\subsection{Style and content module analysis}
\label{subsec:module_study}
In this subsection, we provide extensive analysis of our design choice for the style encoder and the content encoder.
\paragraph{Localized style encoding.} We compare two universal-style encoding strategies to our localized style encoding strategy. First, we train a universal style encoder, which extracts a universal style from the references. EMD, AGIS-Net, and FUNIT employ this scheme. We also develop alternative universal-style encoding strategy with a component-wise style encoder $E_{s,u}$. This alternative encoding utilizes $E_{s,u}$ to extract component-wise features from references; however, the extracted features are directly used without considering the target character. On the other hand, our localized style encoder encodes the character-wise localized style representations using $E_{s,u}$ and factorization modules.

We conduct the ablation study to investigate the effects of different style encoding strategies and summarize the results in Table~\ref{table:abl_style} (the same evaluation setting as Table~\ref{table:main_fewshot}). In Table~\ref{table:abl_style}, we observe that the universal style encoding without $E_{s,u}$ shows comparable style-aware performances (33.6\%) to AGIS-Net (33.3\%) or FUNIT (38.0\%). We further confirm that universal styles by adding the component-wise style encoder $E_{s,u}$ is useful to increase the style-aware metric (33.6\% $\rightarrow$ 52.8\%), and our reorganized localized style representation improves the style-aware metric (33.6\% $\rightarrow$ 72.8\%). The generated samples for each ablation are shown in Figure~\ref{fig:example_ablation}. From these results, we conclude that the proposed localized style representation enables the model to capture diverse local styles, while the universal style encoding fails in fine and local styles.

\begin{table}[t]
\small
\centering
\setlength{\tabcolsep}{4pt}
\begin{tabular}{@{}lccc@{}}
\toprule
Style representation $f_s$      & Acc (S) $\uparrow$    & Acc (C) $\uparrow$    & Acc (Hmean) $\uparrow$  \\ \midrule
AGIS-Net                      & 33.3                  & \textbf{99.7}                  & 49.9 \\
FUNIT                         & 38.0                  & 96.8                  & 54.5 \\ \midrule
Universal without $E_{s,u}$    & 33.6                  & 97.2         & 49.9                    \\
Universal with $E_{s,u}$       & 52.8                  & 95.9                  & 68.1                    \\ \midrule
Localized with $E_{s,u}$      & \textbf{72.8}         & 97.1                  & \textbf{83.2}
            \\ \bottomrule
\end{tabular}
\caption{\small {\bf Impact of localized style representation.} Three different style encoding strategies are evaluated. The universal style encoding without the component-wise style encoder $E_{s,u}$ is defined for each style. The universal style with $E_{s,u}$ is computed by the average of the reference component-wise styles. Our results are shown in the bottom row.}
\label{table:abl_style}
\end{table}

\begin{table}[t]
\small
\centering
\setlength{\tabcolsep}{5pt}
\begin{tabular}{@{}lccccccc@{}}
\toprule
 & \multicolumn{3}{c}{Few-shot} && \multicolumn{3}{c}{Many-shot} \\ 
Accuracies & \multicolumn{1}{c}{S} & \multicolumn{1}{c}{C} & \multicolumn{1}{c}{H} && \multicolumn{1}{c}{S} & \multicolumn{1}{c}{C} & \multicolumn{1}{c}{H} \\ \midrule
DM-Font & 11.1 & 53.0 & 18.4 && 51.8 & 15.0 & 23.2 \\ \midrule
\ours without $E_c$ & 36.3 & 15.4 & 21.7 && 37.8 & 5.1 & 8.9 \\
\ours & \textbf{72.8} & \textbf{97.1} & \textbf{83.2} && \textbf{74.7} & \textbf{96.5} & \textbf{84.2} \\ \bottomrule
\end{tabular}
\caption{\small {\bf Impact of content representation.} We evaluate DM-Font, \ours without content encoder $E_c$, and \ours, in the few-shot (8 references) and many-shot (256 references) scenarios. We report the style-aware accuracy (S), content-aware accuracy (C), and their harmonic mean (H). Note that DM-Font is similar to \ours without $E_c$, but the {\it persistent memory} is used.
}
\label{table:abl_content} 
\end{table}

\paragraph{Content encoding.}
Although the localized style encoding brings impressive improvements in transferring a target style, our localized style encoding strategy has a drawback; it will extract the same feature for characters whose components are identical, but the locations vary. To solve this problem, we employ the content encoder $E_c$ enforced to capture structural information. Here, we examine various content-encoding strategies: \ours without content-encoding, DM-Font ({\it persistent memory} for content encoding), and \ours. When developing LF-font without the content encoder $E_c$, the target glyph is generated with the localized style features alone. DM-Font replaces the content encoder with {\it persistent memory}, which is a learned codebook defined for each component. Note that DM-Font cannot generate unseen reference components; thus we replace unseen component features to the source style features. For removing unexpected effects from this source style replacement strategy, we reported many-shot (256 references) results in addition to few-shot (8 references) results.

In Table~\ref{table:abl_content}, we observe that the content encoder notably enhances overall performance (21.7\% $\rightarrow$ 83.2\% in few-shot harmonic mean accuracy). Since there is no content information, the style encoder of \ours without $E_c$ should encode both style and content information of each component. However, as the style encoder is optimized for modeling local characteristics, it is limited to handle global structures \eg, the positional relationship of components. Besides, because a combination of a component set can be mapped to diverse characters as Figure~\ref{fig:twins}, solely learning localized style features without global structures cannot reconstruct the correct character even though it can capture detailed local styles. Qualitative examples for \ours without the content encoder are in Figure~\ref{fig:example_ablation}.

Similar to the content encoder, the persistent memory strategy proposed by DM-Font, moderately improves the content performance (15.4\% $\rightarrow$ 53.0\%) but shows worse stylization due to the source style replacement strategy. Furthermore, both \ours without $E_c$ and DM-Font suffers from the content performance drop in the many-shot setting. This is because, their style encoders suffer from encoding the complex structures, \eg, relative size or positions, of the unseen styles as shown in Figure~\ref{fig:main_few} (the yellow boxes).

\begin{figure}[t]
    \centering
    \includegraphics[width=\linewidth]{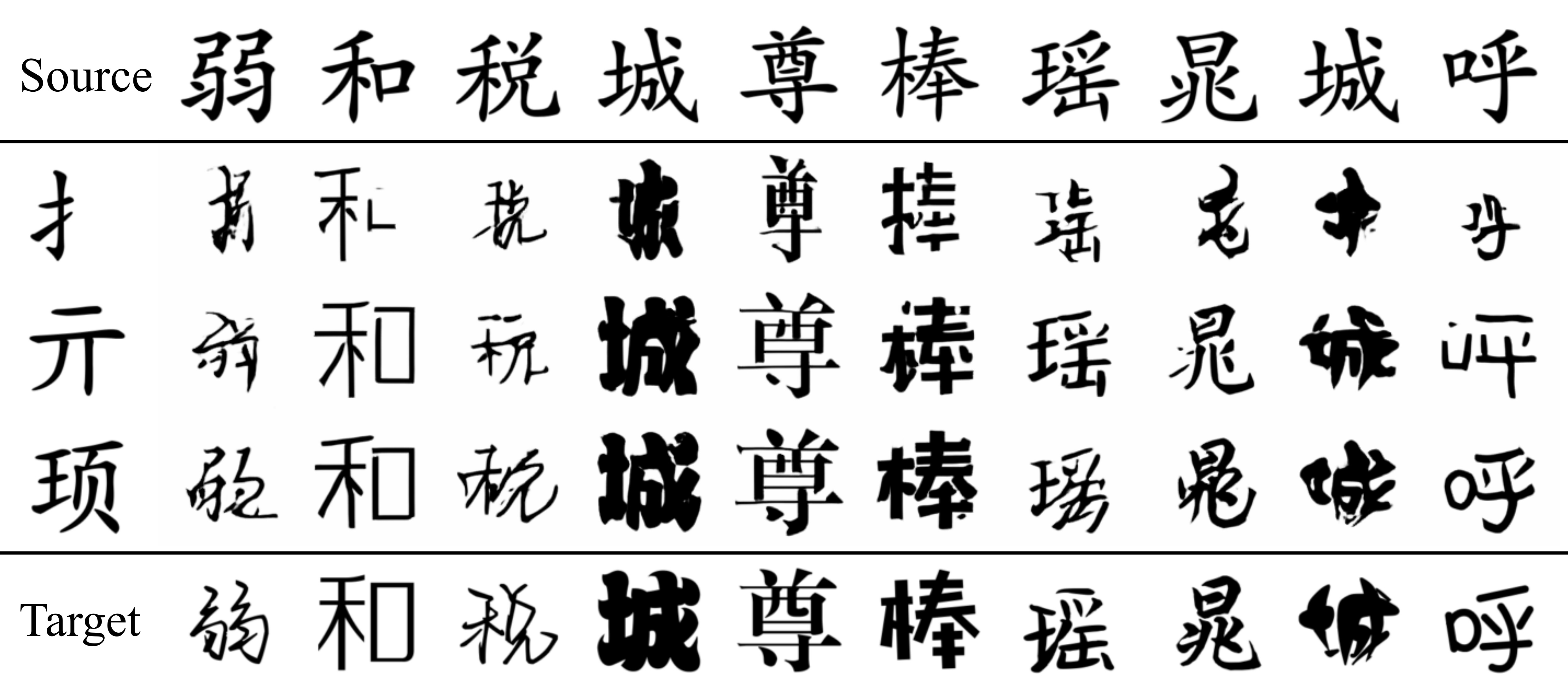}%
    \caption{\small {\bf One-shot generation results.} The reference characters and resultant images are visualized. The top and bottom rows show the source and target images, and the leftmost column shows the single reference used to generate the images in the same row.
    }%
    \label{fig:1shot}
\end{figure}

\paragraph{One-shot generation.} 
We also visualize the extreme case, the one-shot generation by \ours, in Figure~\ref{fig:1shot}. We observe that when the reference glyph is too simple to extract solid component-wise features (the second row in Figure~\ref{fig:1shot}), the generated images show poor visual quality. Note that this might be the problem of style factors, not $E_c$, because the same content factors and content features (from $E_c$) are used for successfully generating other samples. Hence, we can conclude that the reference selection is critical to the visual quality and that the reference having rich local details is advantageous for high-quality generation.

\section{Conclusion}

Our novel few-shot font generation method, named \ours, produces complex glyphs that preserve the local detail styles by introducing character-wisely defined style representations. Furthermore, we propose the factorization modules to reconstruct the entire character-wise style representations from a few reference images. It enables us to reorganize the seen character-wise style representations to the unseen character-wise style representations by disentangling character-wise style representations into style and component factors. In the experiments, \ours outperforms state-of-the-art few-shot font generation methods in various evaluation metrics, particularly in {\it style-aware} benchmarks. Our extensive analysis of our design choice supports that our framework effectively disentangles content and style representations, resulting in the high-quality generated samples with only a few references, \eg, 8.

\section*{Acknowledgement}
We thank Junsuk Choe, Dongyoon Han, Chan Young Park for the feedback. NAVER Smart Machine Learning (NSML)~\cite{nsml} has been used for experiments. This research was supported by the Basic Science Research Program through the National Research Foundation of Korea (NRF) funded by the MSIP (NRF-2019R1A2C2006123, 2020R1A4A1016619), and by Institute of Information \& Communications Technology Planning \& Evaluation (IITP) grant funded by the Korea government (MSIT) (2020-0-01361, Artificial Intelligence Graduate School Program (YONSEI UNIVERSITY)). This work was also supported by the Korea Medical Device Development Fund grant funded by the Korea government (the Ministry of Science and ICT, the Ministry of Trade, Industry and Energy, the Ministry of Health \& Welfare, Republic of Korea, the Ministry of Food and Drug Safety) (Project Number:  202011D06  )

{\small
\bibliography{reference}
}

\appendix
\section{Additional Experimental Results}

\subsection{Reference image samples}

\begin{figure}[ht]
    \centering
    \includegraphics[width=\linewidth]{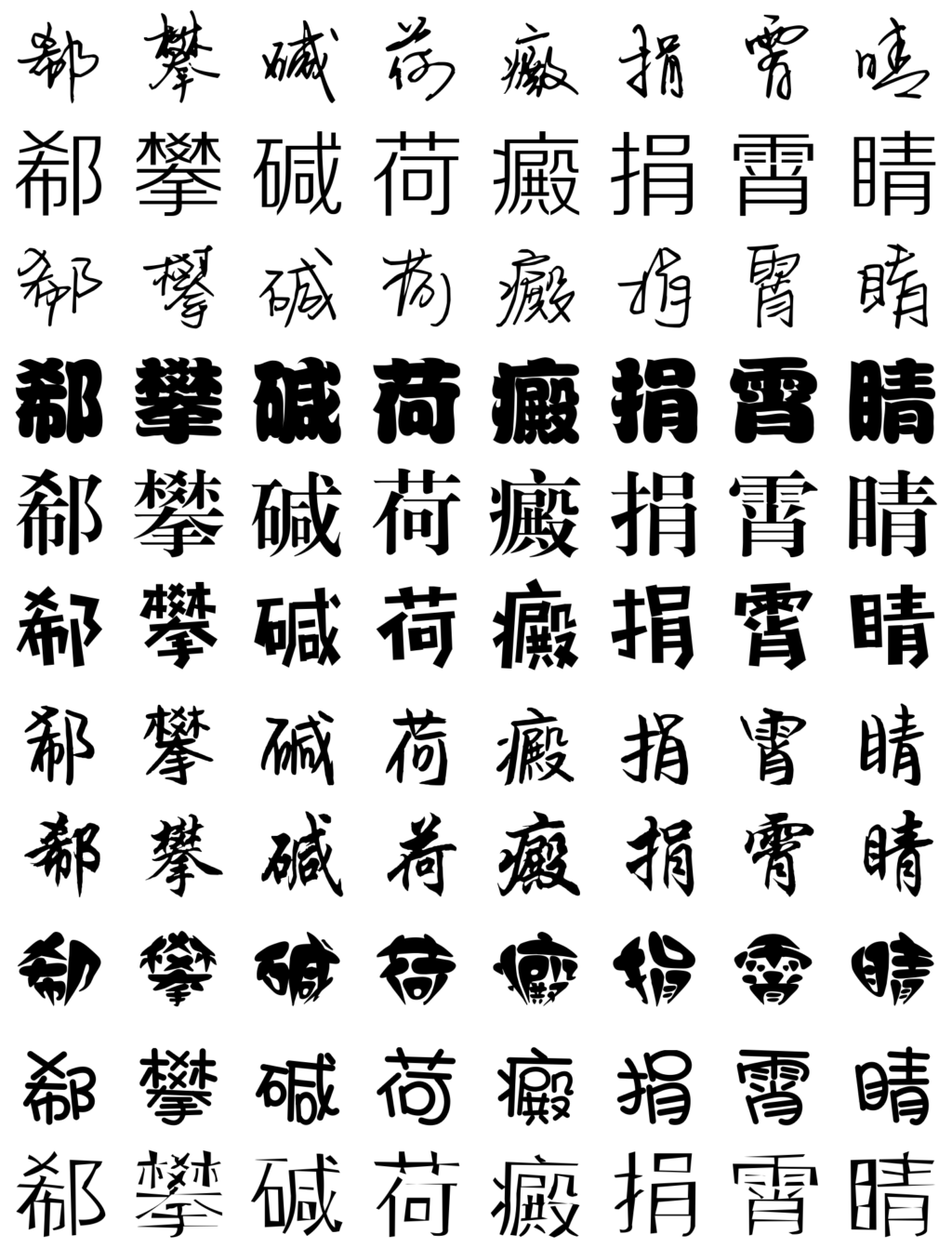}%
    \caption{\small {\bf Reference images with target styles.} We visualize the eight reference samples per style used in the Figure~4. Each row corresponds to two columns of Figure~4, in the same order.
    }%
    \label{fig:style_refs}
\end{figure}

\subsection{Parameter search}
We provide ablation study on the effects of the consistency loss $\mathcal L_{consist}$, the component classification loss $\mathcal L_{cls}$ and the factor dimension $k$. Table~\ref{table:abl_loss} shows that both $\mathcal L_{consist}$ and $\mathcal L_{cls}$ enhance the overall performance. $\mathcal L_{cls}$ particularly improves the style-aware accuracy (44.5 $\rightarrow$ 69.3). We also report the performances with different factor dimension $k$ in Table~\ref{table:abl_k}. The best parameter $k=8$ is used in this paper.

\begin{table}[ht!]
\small
\centering
\begin{tabular}{@{}ccccc@{}}
\toprule
$\mathcal L_{consist}$ & $\mathcal L_{cls}$ & Acc (S) $\uparrow$ & Acc (C) $\uparrow$ & Acc (Hmean) $\uparrow$ \\ \midrule
\nomark  & \nomark   & 44.5             & 76.3          & 56.2   \\
\yesmark & \nomark   & 47.2             & 88.6          & 61.6   \\
\nomark  & \yesmark  & 69.3             & \textbf{97.2} & 81.1              \\ \midrule
\yesmark & \yesmark  & \textbf{72.8}    & 97.1          & \textbf{83.2}     \\ \bottomrule
\end{tabular}
\caption{\small {\bf Impact of objective functions.}
We report the accuracies of the different combinations of consistency loss $L_{consist}$ and the component-classification loss $L_{cls}$.
Our design choice is the bottom row, which shows the best overall performance.}
\label{table:abl_loss}
\end{table}%
\begin{table}[ht!]
\small
\centering
\begin{tabular}{@{}cccc@{}}
\toprule
$k$                 & Acc (S) $\uparrow$ & Acc (C) $\uparrow$ & Acc (Hmean) $\uparrow$          \\ \midrule
4                       & 71.0          & 98.0              &82.3               \\
6                       & 72.0          & \textbf{98.0}              &83.0               \\
8$^\dagger$             & \textbf{72.8} & 97.1     & \textbf{83.2}     \\
10                      & 71.4          & 97.5              & 82.4              \\ \bottomrule
\end{tabular}
\caption{\small {\bf Factor size study.} The results on varying factor sizes are reported. $k$ denotes the factor size. $^\dagger$ used in the remaining experiments.} 
\label{table:abl_k}
\end{table}

\begin{figure*}[ht!]
    \centering
    \includegraphics[width=\linewidth]{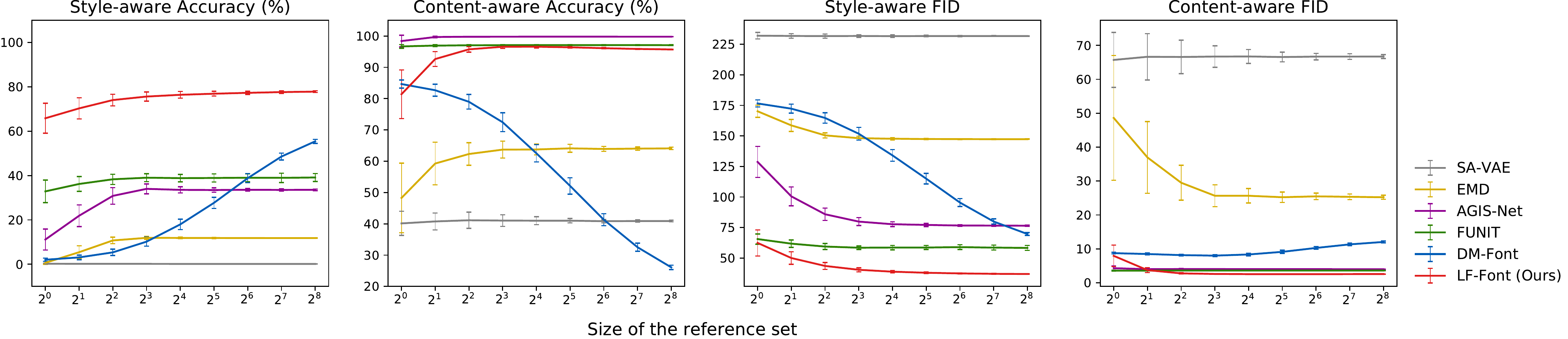}%
    \caption{\small{\bf Performance changes by varying size of the reference set.} We report how the performances of each model are affected by the size of the reference set $\mathcal X_r$. The style-, content-aware performances are evaluated with generating seen characters and each recorded in two metrics, accuracy (higher is better) and FID (lower is better). Each graph shows the average performance as a line and errors as an errorbar.
}%
    \label{fig:numref_performance}
\end{figure*}

\begin{figure*}[ht!]
    \centering
    \includegraphics[width=\linewidth]{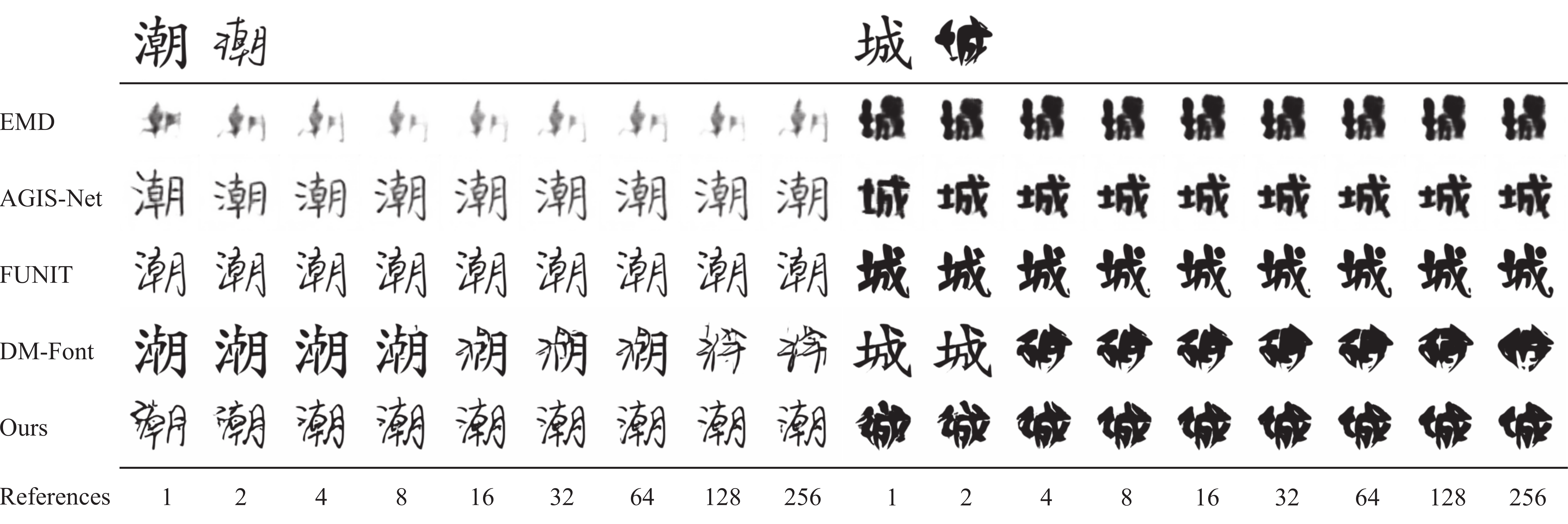}%
    \caption{\small {\bf Generated samples by varying reference set size.} Each row shows the generated samples by each model. The source and target glyphs are displayed in the top row.}%
    \label{fig:nref_sample}
\end{figure*}

\subsection{Reference size study}
We report the performances of few-shot methods by varying the size of the reference set in Figure~\ref{fig:numref_performance}. \ours remarkably outperforms other methods in style-aware metrics. Furthermore, \ours in one-shot shows better results than others in many-shot in terms of the style-aware metrics.
Compared to \ours, FUNIT but AGIS-Net show stable content-aware performances in the low reference regime, and the generated samples are less stylized than \ours as shown in Figure~\ref{fig:nref_sample}.

In Figure~\ref{fig:numref_performance}, we observe that most methods show better performance with more references, except DM-Font; although DM-Font is designed for many-shot, the overall performance rather drops as the references increase in Chinese. As discussed in \S~4.4, it is because the absence of the content encoder damages in capturing the complex glyph structures such as Chinese characters, while DM-Font is intended for complete compositional scripts, \eg Korean.

\subsection{Style interpolation}

\begin{figure}[ht!]
    \centering
    \includegraphics[width=\linewidth]{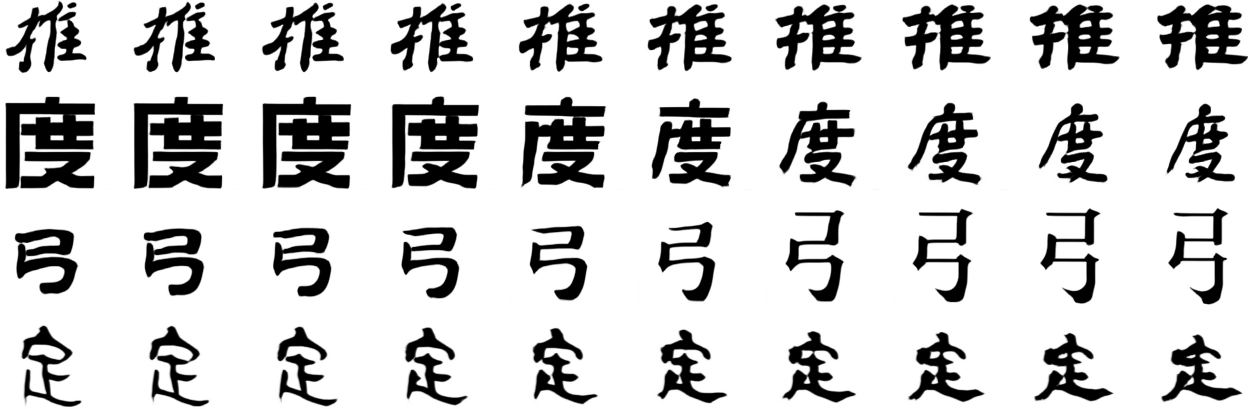}%
    \caption{\small {\bf Style interpolation.} Generated glyphs in each row correspond to identical content representation and component factors, but differ only in the style factors. 
    }%
    \label{fig:style_mix}
\end{figure}

To show that our style representations are semantically meaningful, we provide the style interpolation results in Figure~\ref{fig:style_mix}. \ours shows well-interpolated local features such as diverse component size, serif-ness, or thickness.

\begin{figure*}[t]
    \centering
    \includegraphics[width=\linewidth]{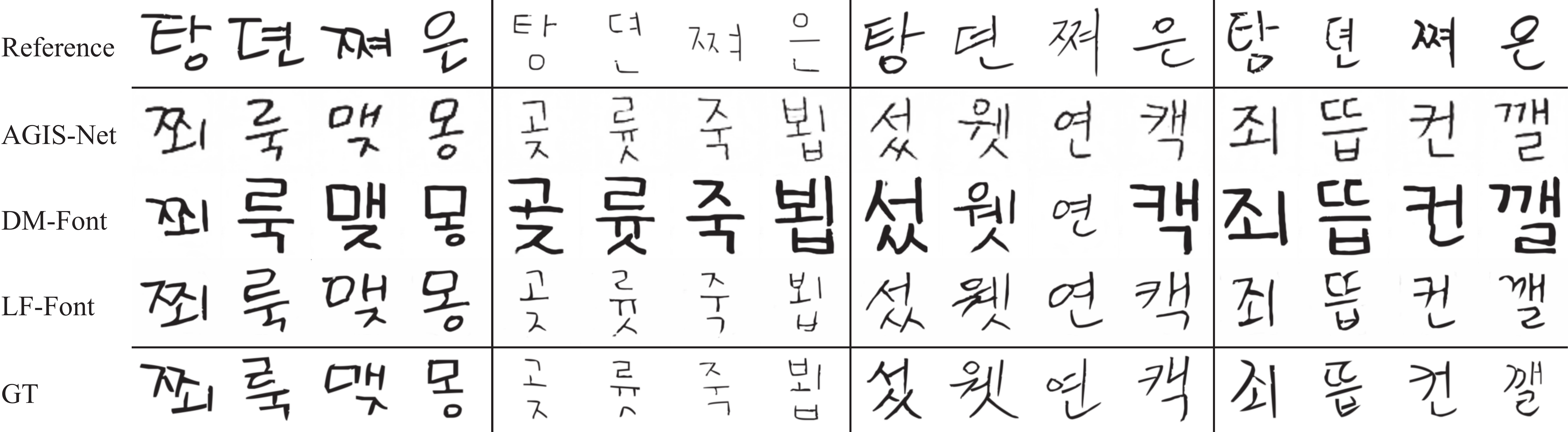}%
    \caption{\small {\bf Korean few-shot generation samples.} The generated samples by each model and the ground truth glyphs are shown. The samples are generated with four reference images which are shown in the top row.}%
    \label{fig:kor_result}
\end{figure*}

\subsection{Few-shot Korean generation}

We report the Korean few-shot generation results with four reference glyphs in Table~\ref{table:app_kor}.
We compare LF-Font to two state-of-the-art Korean few-shot generation methods, AGIS-Net~\cite{gao2019agisnet} and DM-Font~\cite{cha2020dmfont}.
We use the same evaluation classifiers used in \citet{cha2020dmfont}.
As Chinese experiments, we report the average of LPIPS, FID and accuracies of ten different runs with different reference selection to avoid randomness by the references.
In the table, we observe that LF-Font outperforms DM-Font and AGIS-Net in overall metrics, particularly in {\it style-aware} metrics.
The visual examples are shown in Figure~\ref{fig:kor_result}.
\begin{table*}[ht!]
\small
\centering
\begin{tabular}{@{}lccccccc@{}}
\toprule
                    & LPIPS $\downarrow$ & Acc (S) $\uparrow$ & Acc (C) $\uparrow$ & Acc (Hmean) $\uparrow$ & FID (S) $\downarrow$ & FID (C) $\downarrow$ & FID (Hmean) $\downarrow$ \\ \midrule
AGIS-Net (TOG'19)   & 0.188        & 3.9 & 97.5 & 7.5 & 108.1 & 7.8 & 14.5 \\
DM-Font (ECCV'20)   & 0.266        & 3.4 & 96.3 & 6.5 & 126.3 & 19.0 & 33.0 \\
LF-Font (proposed)  & \textbf{0.145} & \textbf{41.6} & \textbf{98.4} & \textbf{58.5} & \textbf{47.2} & \textbf{4.9} & \textbf{8.9} \\ \bottomrule
\end{tabular}
\caption{\small {\bf Performance comparison on few-shot Korean font generation scenario.} We report LPIPS, FID and accuracy measures for AGIS-Net, DM-Font and LF-Font. All numbers are average of 10 runs with different reference glyphs.}
\label{table:app_kor}
\end{table*}

\section{Implementation Details}

\subsection{Network architecture details}
The component-wise style encoder $E_{s,u}$ consists of five modules; convolution, residual, component-conditional, global-context~\cite{cao2019gcnet}, and convolutional block attention (CBAM)~\cite{woo2018cbam}.
Our component-conditional block is implemented as a set of channel-wise biases and each bias value corresponds to each component. We employ the global-context block and CBAM to enable the network to capture correct components. The content encoder $E_c$ and the generator $G$ consist of convolution and residual blocks. More detailed architecture is in our code.

\subsection{\ours implementation details}
We use Adam~\cite{adam} optimizer with learning rate 0.0008 for the discriminator and 0.0002 for the remaining modules. 
We train the model in two-phase for stability. In the first phase, we train the model without factorization modules during 800k iterations for Chinese and 200k iterations for Korean. In this phase, $\lambda_{consist}$ is set to $0.0$ and component-wise style features from $E_{s,u}$ are used to generate target glyph and classified by component-wise classifier $Cls$. 
After enough number of iterations, we add the factorization modules to the model and jointly train all modules with the full objective function for 50k iterations. All the component-wise style features used in the first phase are replaced to reconstructed component-wise style features from style and component factors.
The minibatch for training are set differently in each phase. Since the model cannot deal with component-wise style features not in the reference set without factorization modules, the images in the reference set and target glyph share the same style in the first phase.
In the second phase, the images in the reference set have various styles, and the target glyph has one of them. We let the model reconstruct the reference images to prevent the model from losing the first phase's performance.

\subsection{DM-Font modification details}
As DM-Font cannot generate Chinese characters, we modified the structure of DM-Font in our Chinese few-shot generation experiments. Since Chinese characters are not decomposed into the same numbers of components, we modified the multi-head structure to a component-condition structure the same as \ours and used the averaged component-wise style features as an input of the decoder. We also changed its attention blocks to CBAM and eliminated the hourglass blocks in its decoder to stabilize the training.
For Korean few-shot generation experiments, we used the official DM-Font model and the trained weight.

\subsection{Evaluation classifier details}
We set CutMix probability and the CutMix beta to 0.5 and 0.5, respectively. We also employ AdamP~\cite{heo2020adamp} optimizer, and the batch size, the learning rate, the number of epochs are set to 64, 0.0002, 20.

\end{document}